\documentclass[journal]{IEEEtran}
\usepackage{microtype}
\usepackage{lmodern}
\usepackage{subfigure}
\usepackage{booktabs}
\usepackage{url}

\usepackage[style=base,font=small]{caption}

\title{\LARGE \bf
The Text-Based Adventure AI Competition
}

\author{Timothy Atkinson$^{1*}$, Hendrik Baier$^{2*}$, Tara Copplestone$^{1}$,
Sam Devlin$^{2}$ and Jerry Swan$^{1}$ 
\thanks{$^{1}$ Department of Computer Science,
        University of York, UK.}%
\thanks{$^{2}$ Digital Creativity Labs, University of York, UK.}%
\thanks{$^{*}$ These authors contributed equally to this work.}%
\thanks{Corresponding authors: {\tt\small tja511@york.ac.uk}, {\tt\small hendrik.baier@gmail.com }}
}

\usepackage{listings}
\begin{document}

\maketitle
\thispagestyle{empty}
\pagestyle{empty}


\begin{abstract}
In 2016, 2017, and 2018 at the IEEE Conference on Computational Intelligence in Games, the authors of this paper ran a competition for agents that can play classic text-based adventure games. This competition fills a gap in existing game AI competitions that have typically focussed on traditional card/board games or modern video games with graphical interfaces. By providing a platform for evaluating agents in text-based adventures, the competition provides a novel benchmark for game AI with unique challenges for natural language understanding and generation. This paper summarises the three competitions ran in 2016, 2017, and 2018 (including details of open source implementations of both the competition framework and our competitors) and presents the results of an improved evaluation of these competitors across 20 games. 
\end{abstract}


\section{INTRODUCTION\label{sec:introduction}}
Before the widespread availability of graphical displays, text adventures were one of the few game genres that owed their existence solely to computing. The first text adventure was \textsc{Colossal Cave} (also known simply as \textsc{Adventure}), written in 1976 by Will Crowther for the PDP-10 mainframe \cite{Montfort:2004:TLP:940352}. With the advent of home computing in the late 1970s, \textsc{Colossal Cave} and other games such as \textsc{Zork} were enjoyed by many. The majority of early text adventures used a narration-action loop that accepted simple commands of the general form {\tt VERB} or {\tt VERB NOUN} (e.g.\ {\tt `look'}, {\tt `go west'}, {\tt `take box'}) via console input. In response to such commands, the programs provided a description of the immediate environment, e.g.\ 

{\tt `You are in an open field on the west side of a white house with a boarded front door. There is a small mailbox here.'}

Early adventures typically involved exploration and treasure hunting, but more sophisticated narratives emerged in the 1980s (e.g. in the \textsc{Infocom} range of games). An active ``Interactive Fiction'' community still exists, using powerful natural language authoring tools such as \textsc{Inform} \cite{Nelson00theinform} to create new and diverse titles.

Despite the continued existence of this community, one might ask what a competition concerned with text-based games has to offer, given that there are numerous competitions involving modern graphics-based games. The underlying motivation can be traced back to an early divergence between AI philosophy and practice. John McCarthy proposed the well-known ``Monkey and Bananas Problem'' in 1963 \cite{mccarthy1963situations}: given a room containing a chair, a stick and a bunch of bananas hanging on a hook, the monkey's task is to find a sequence of actions that results in acquiring the bananas. McCarthy made a key distinction between the \emph{physical feasibility} of the task (i.e.\ is there a physically realisable sequence of actions that achieves the goal?) and its \emph{epistemic feasibility} (i.e.\ can the knowledge that a particular action even \emph{exists} be efficiently derived?) \cite{McCarthy:1977:EPA:1622943.1623044}. We claim that recent game AI competitions have tended to de-emphasise the epistemic aspect, instead working in domains that are strongly \emph{operationalised}. What this means is that the set of actions instantaneously available have been constrained to be knowable in advance. Of course, the space of plans for lengthy \emph{sequences} of actions is still combinatorially huge, but the key question of how to derive and represent knowledge about an uncertain world is circumvented. 

The hypothesis motivating the Text-Based Adventure AI competition is that the determination of relevant \emph{affordances} (i.e.\ the set of behaviors that are possible in a given situation \cite{gibson1968senses}) from a non-trivial environment is likely to require more than a good choice of credit assignment strategy, even if the latter would suffice in an operationalised domain. This then re-emphasises the following research questions:
\begin{itemize}
\item In the absence of operationalisation, what are the minimum \emph{priors} (in terms of domain knowledge) that are required for success (even for a ``toy'' domain such as ``Monkey and Bananas'')?
\item How do model-free and model-based approaches compare? How can any difficulties with the former in this domain better inform such approaches in general?
\end{itemize}

In operationalised environments, the goal is to learn a policy mapping from game states to actions, where the set of available actions is predetermined.
In a non-operationalised environment, the set of available actions first has to be generated as a function of previously encountered game states. A policy mapping that works with these generated actions can then be learned subsequently.
This function space is in general vastly larger than in the operationalised case.

It could be argued that text adventure domains are also ``operationalised'', since the space of possible inputs is discrete. However, this does not take into account the fact that \emph{effective} action requires a mapping from game output space to player input space for which random input text will be vastly less likely to have any effect than randomly pressing buttons in a first-person shooter, or randomly making legal moves in chess.

This article is structured as follows: Section \ref{sec:relatedwork} discusses related work on natural language processing in game-playing. Section \ref{sec:competitionframework} describes the competition framework. Section \ref{sec:competitors} outlines the agents submitted to the 2016, 2017, and 2018 competitions. Section \ref{sec:evaluationmethodology} describes an improved evaluation methodology based on the experience gained from the 2016 and 2017 competitions. Section \ref{sec:results} presents new results from applying this methodology to the existing agents, which we hope can serve as a baseline for future research, and Section \ref{sec:conclusion} concludes.

\section{RELATED WORK\label{sec:relatedwork}}
The essential task of a competitor in the Text-Based Adventure AI Competition is to create an agent that can act effectively in a partially-observable environment, perceived via a number of short natural language descriptions of the agent's surroundings. Historically, work in the general area of natural language processing and narrative representation has mirrored the overall tendency for AI methods to move from symbolist to data-intensive methods such as Neural Networks, Reinforcement Learning or Monte Carlo Tree Search. In the 1970s and 1980s there was significant interest in symbolic representations of natural language and narrative, via semantic nets, frame systems and scripts, using approaches such as Case-Based Reasoning \cite{schank:77a} and Abstraction Units \cite{LEHNERT1981293}. More recently, there has been a tendency for a directly symbolist approach to Natural Language Processing to be eclipsed by more overtly numerical methods, including ``Bag of Words'' based approaches such as Latent Semantic Analysis \cite{ASI:ASI1}, with the goals of approaches such as \cite{LEHNERT1981293} being revisited via contemporary corpus-based techniques \cite{Goyal:2010:APP:1870658.1870666}.

With specific reference to work on games, Branavan et al. \cite{DBLP:conf/ijcai/BranavanSB11} applied Monte Carlo Tree Search to the strategy game \textsc{Civilization II}. The value function was approximated via a neural net and included linguistic features extracted from the game manual, providing a significant improvement against the game's built-in opponent. Narrowing our focus further to Interactive Fiction, AI can play many roles in this genre of games \cite{riedl2012interactive} with this competition's focus being specifically on game playing AI. Previous work has often attempted to reduce the challenge of text-based games in order to make them more feasible domains for existing AI methods - such as for example by making the game output its environment descriptions not in natural language, but in first-order logic that can be directly stored and processed by an agent \cite{hlubockyknowledge,amirlearning}. Prior to our first competition in 2016, the previous work of greatest relevance in playing natural-language games was by Narasimhan et al \cite{DBLP:journals/corr/NarasimhanKB15}, in which Deep Reinforcement Learning was used to jointly learn state representations and action policies in two Multi-User Dungeons (MUDs) - a specific type of text-based adventure game. The approach was demonstrated to outperform the use of both bag-of-words and bag-of-bigrams for state representations. This paper and early discussions with the first two authors helped shape the first competition we ran in 2016, which attempted to engage the game AI community more broadly in the challenging topic of text-based adventure AI by providing an accessible framework and regular schedule for agent evaluation via competitions.

\section{COMPETITION FRAMEWORK} \label{sec:competitionframework}
Our framework for evaluating a software agent's ability to play text-based adventures\footnote{The Text-Based Adventure AI framework is available at \url{https://github.com/Atkrye/IEEE-CIG-Text-Adventurer-Competition}} is based on the \textsc{ZPlet} \cite{zplet} Java interpreter for the widely-used Z-Machine format \cite{zmachinestandardsdoc}, created by Infocom in 1979. The framework is defined in terms of a traditional agent-based perspective \cite{Russell:2003:AIM:773294}. The console input and Z-Machine output text of \textsc{ZPlet} are redirected to an \emph{Agent} interface, to be implemented by competitors. This interface consists of a single \emph{action} method with a string argument. At each turn, the narrative text (typically at most a paragraph in length) that would otherwise be presented to the human player is provided as this argument. For example:

{\tt `You are standing at the end of a road before a small brick building. Around you is a forest. A small stream flows out of the building and down a gully'}. 

\normalsize The \emph{action} method then returns a string describing the action that the agent wishes to perform. A default \emph{RandomAgent} is provided, with an \emph{action} implementation that simply ignores the input text and chooses uniformly from 8 basic commands: {\tt `north'}, {\tt`south'}, {\tt`east'}, {\tt`west'}, {\tt`verbose'} (switching some games to a mode where they always output full descriptions of locations even if the agent has been there before), {\tt`take all'}, {\tt`yes'}, and {\tt`no'}.


By default, agents can be developed in Java\footnote{A tutorial for getting started with developing Java agents is available at: \url{http://atkrye.github.io/IEEE-CIG-Text-Adventurer-Competition/tutorial-java/}}, but to facilitate the development and implementation of agents in other programming languages an additional agent, \emph{IOAgent}, is provided. This agent's \emph{action} method forwards the narrative text to the application's output stream and then returns the next line of text from the application's input stream. This allows entrants to interface external agents by starting the implementation with the \emph{IOAgent} and then interacting with that process's IO. A random agent implemented  in Python 3.5 is provided to demonstrate how this can be implemented. 

Additionally, there are 3 predefined actions which are meant to help train an agent and may be accessed by the \emph{IOAgent} through hard-coded string equivalents. These are \emph{Quit}, which quits the running game; \emph{Restart}, which restarts both the running game and the current agent; and \emph{SoftRestart}, which restarts the running game but not the current agent. 


\section{COMPETITORS\label{sec:competitors}}
The first Text-Based Adventure AI Competition was announced on 15 May 2016 and ran later that year at the IEEE Conference on Computational Intelligence in Games after closing for entries on 31 August 2016. The competition ran again in 2017, officially being announced on 8 March 2017 and closing for entries on 18 July 2017. In 2018, the competition was officially announced on 3 May 2018 and the closing for entries was 20 July 2018. In the following subsections we will summarise the four agents submitted to these competitions, three of which have been made available as open source by the respective participants. Future papers based on our framework can use these agents as baselines for evaluation and comparison.
%
%
%
%

\subsection{\textsc{BYU\-Agent 2016}}

\textsc{BYU\-Agent 2016}\footnote{The BYU\-Agent 2016 \cite{byu2016} is available open source at: \url{https://github.com/danielricks/BYU-Agent-2016}} \cite{byu2016} interacts with a text-based adventure by generating commands combining nouns extracted from the game text with verbs drawn from a predefined set. More complex commands, such as propositional structures, may also be generated. Generated commands are attempted exhaustively and those commands which successfully produce a change in the game environment are stored. When the agent encounters a game environment where generated commands fail to produce a change, these stored commands may then be attempted, effectively implementing a simple form of one-shot learning. 

The set of verbs the agent uses is drawn from the Wikipedia text corpus. The 1000 most commonly-appearing verbs were extracted, and then filtered by human play-testers according to their usefulness on a variety of text-based adventure games in order to produce a smaller set of approximately 100 verbs. A small number of additional human-selected verbs, including basic navigational commands, are also included. 

Commands are generated using the word2vec \cite{DBLP:journals/corr/abs-1301-3781} algorithm to produce vector embeddings of verbs and the nouns extracted from the game text. An ``affordance vector'', derived as the average vector difference in a known set of verb-noun pairs, is used to find a set of verbs which ``match'' a given noun. Additional hard-coded behaviour, such as periodic \texttt{`look'} and \texttt{`inventory'} commands to observe the game state, and \texttt{`get all'} whenever entering a new location, aids the agent in interacting with the game.

\subsection{\textsc{Golovin} (2017)}

The \textsc{Golovin}\footnote{The Golovin agent \cite{golovin} is available open source at: \url{https://github.com/Kostero/text_rpg_ai}} agent \cite{golovin} uses command generators to propose a non-empty set of commands for a given game environment by inserting nouns taken from the game's narrative text into ``command patterns'' - a set of 250,000 verb phrases extracted from various game walkthroughs, tutorials, and raw narrative text. To do this, a word2vec model trained on 3000 fantasy books is used to propose synonyms for each noun using n-best cosine similarity. Commands are then proposed by finding command patterns containing these synonyms and replacing that synonym with the original noun. Each generated command is associated with a weight, consisting of multiple factors including the cosine similarity between the noun and its synonym's word2vec vector representation and a value given by a LSTM neural network operating on words \cite{DBLP:journals/jmlr/BengioDVJ03}. Finally, a roulette wheel selection (based on the commands' associated weights) is used to choose a command from the generated set.

When a command is attempted and the game description remains the same, that command is assumed to have failed and is blacklisted for that game location until the agent observes a change in its inventory. 

Five distinct command generators exist for different purposes. Each of these generators is fired, in this order, until a non-empty set of commands is proposed:
\begin{enumerate}
\item Battle mode. This command generator, specifically designed to aid the agent in combat scenarios, is limited to a subset of around 70 ``fighting'' command patterns containing one of the verbs \texttt{`attack'}, \texttt{`kill'}, \texttt{`fight'}, \texttt{`shoot'} or \texttt{`punch'}. This generator does not blacklist ``failed'' commands, as combat may require multiple iterations, and is only fired if a ``fighting''  command has been used previously. 
\item Gathering items. This command generator is fired whenever the agent enters a new area, proposing \texttt{`take'} commands for nouns in the area's narrative text.
\item Inventory commands. Once a noun has been successfully taken by the gathering items generator, this generator may fire, generating commands under the usual approach using only that noun's synonyms.
\item General actions. This command generator proposes commands in the usual approach using nouns from the game environment.
\item Exploration. A small fixed set of movement commands are proposed when an area has been exhausted by other command generators. Once the agent moves to a new area, a map graph storing locations and the commands which move between them is updated. If an area has unexplored directions, then those directions are proposed at random. Otherwise a route to a ``promising destination'', measured by its distance away and the proportion of possible commands left unattempted, is attempted. 
\end{enumerate}


\subsection{\textsc{CARL} (\textsc{BYU\-Agent 2017})}

The CARL agent uses affordance detection to suggest commands based on objects observed in a game's narrative text. Additionally, when an action is perceived to change the game state, that action is stored in memory so that it can be repeated during subsequent encounters with that state.

States are identified and stored by converting the individual sentences of the narrative text into skip-thought vectors \cite{DBLP:journals/corr/KirosZSZTUF15}. The vector of each sentence is then classified as either state-information or not based on its proximity to the vector representations of a set of labelled example sentences. Those sentences which are identified as state-information are concatenated and hashed to create a unique identifier of the current state for the action recollection strategy described above. 

Commands are then generated by first extracting nouns from those sentences identified as state-information. The vector representations of these nouns, generated by a word2vec model, trained on the Wikipedia text corpus, are manipulated using linear algebra to identify likely ``matching'' verbs. The best candidate verb-noun pairs are attempted first, with the search broadening to less ``well-matching'' pairs should these fail to change the game's state. A complementary algorithm also attempts to generate prepositional combinations. 

\subsection{\textsc{NAIL} (2018)}

The NAIL (``Navigate Acquire Interact Learn'') agent\footnote{The NAIL agent is available open source at: \url{http://aka.ms/nail}} consists of multiple independent modules which compete for control of the agent. Each module has a specific purpose; the main modules are the \textit{Examiner}, \textit{Interactor} and \textit{Navigator} modules. These modules are responsible for identifying relevant objects in the current location, interacting with identified objects, and navigating to a new location, respectively. Additionally, there are further modules including specific modules for yes-no questions and for the acquisition of objects. Each module observes changes in the game and in each step of the game reports how `eager' it is to assume control of the agent, with the most eager module in any step gaining full control.

Any change in the game's state is used to update a knowledge graph, which all modules have access to. This knowledge graph tracks known objects, interactions, locations and connections between locations, and serves as a compact representation of the world state. Additionally, the knowledge graph stores all previously attempted interactions to avoid retrying failed interactions. 

Most modules use pre-defined sets of common commands to interact with the game. The \textit{Interactor} module deviates here, constructing verb-noun phrases to interact with objects observed in the game. This module employs a LM-Based language model
to assign probabilities to different verb-object combinations, the most promising of which are then executed. When there are no promising verb-object combinations, the \textit{Interactor} falls back to a predefined set of verbs which are attempted in combination with observed objects from the game.

The NAIL agent uses a \textit{validity detector} to determine whether its actions are having an effect on the game. A word-embedding-based text classifier \cite{JoulinGBM16} is used to establish whether a game's response to a given action either ``failed'' (had no effect) or ``succeeded'' (had an effect). Validity detection is needed e.g. to decide whether an object is relevant in the \textit{Examiner} module, and to prevent incorrect updates to the knowledge graph when an action ``failed'.

\section{EVALUATION METHODOLOGY\label{sec:evaluationmethodology}}

The agents submitted to the 2016 and 2017 competitions were evaluated on a single game developed specifically for the competition in order to provide a gradually increasing level of difficulty. The game was written by a game designer without a formal education in AI, with the intention of creating a game unbiased towards any existing approaches to game playing AI. The winner in both years was \textsc{BYU\-Agent 2016}.

In hindsight, we realise the evaluation via a single game was flawed. Due to the mechanics of text-based adventures, if an agent is unable to solve a single puzzle, it is often unable to progress further in the game. This became obvious in the 2017 competition, where both the \textsc{Golovin} and CARL agents failed to score a single point because they could not solve the first puzzle. We had assumed the first puzzle to be simple, but our assessment of the complexity of text-based puzzles was entirely subjective. Moreover, the relative performance of agents in this competition compared to their performance as reported in the competitors' own publications \cite{byu2016,golovin} suggest that our game was biased towards certain types of agents (e.g. \textsc{BYU\-Agent 2016}). 


This motivated the need for an evaluation across multiple games in order to fairly judge agents capable of general text-based adventure game playing\footnote{The authors of \textsc{Golovin} and the BYU agents themselves seem to have already agreed on a training set of 50 text-based adventure games for algorithm development and testing \cite{byu2016,golovin}.}. The advantages of AIs that can play multiple games instead of a single game have frequently been promoted by the research community \cite{DBLP:conf/aaai/LiebanaSTSL16,genesereth2005general}, and there is a growing trend towards recognising the advantages of evaluating agents on commercial games not specifically designed to challenge AIs as well \cite{bellemare2013arcade,togelius2015ai}. Specifically, by evaluating on multiple games instead of a single test game, we avoid the tendency of competitors to overfit to the single test game; creating a contribution to that specific game instead of to AI research more broadly. Furthermore, the use of commercial games avoids bias in the game design towards specific AI methods the game designer may wish to promote if the game is created specifically to test AI. By instead using games created for human gameplay, this bias is presumed to be removed or at least averaged out over multiple games.

In the 2018 competition, we therefore switched to a more general evaluation framework using a test set of 20 different text adventures, which we downloaded from the web. The games were written in different styles by different authors, and some had been previously either commercially released, or submitted to interactive fiction writing competitions. There is no overlap between our test set and the training set used by \cite{byu2016,golovin}. Unlike some of the games in this training set, all games in our test set provide a numerical score that allows for a fine-grained measurement of game playing performance. The precise actions for which a player gets rewarded with points, such as the successful solving of puzzles or winning of fights, are determined by the authors of the individual games. We expect that this removes the biases mentioned above -- both regarding our own style of writing, as well as our own style of scoring text-based adventure games. In addition, agents that fail at solving the first puzzle of any given game will still be able to make progress on other games. Agents can be evaluated on their performance given a fixed number of game steps (calls to the \emph{action} method) per game, and their evaluation can be averaged over multiple runs with different random seeds in order to reduce the impact of run-to-run variance.

The following section presents a novel evaluation of all agents submitted to previous competitions using our new framework, with the aim of more accurately determining which agent is currently the best general text adventure game playing AI. We evaluated \textsc{BYU\-Agent 2016}, \textsc{Golovin}, CARL (\textsc{BYU\-Agent 2017}), and \textsc{NAIL} on each of the 20 test games for 1000 game steps. This is to make our results comparable to those in \cite{byu2016,golovin} with the same number of steps. The results reported here are averages over 10 such runs for \textsc{Golovin} and \textsc{NAIL}; unfortunately, we were only able to do one run for each of the BYU agents (discussion below). In order to give an indication of how agent performance can scale with the number of game steps, NAIL and \textsc{Golovin} were additionally tested in 10 runs with 100 steps each, and \textsc{Golovin} was also tested in 10 runs with 10,000 steps each on all 20 games.

The evaluation metrics are the average percentage of points an agent achieved over all games and test runs, and the percentage of games in which an agent achieved any points, averaged over test runs. The first metric is our primary evaluation metric, and is expressed as an average percentage instead of an average number of points due to the large differences in the maximum number of points achievable in each game. 1 point out of a maximum of 10 should count for more than 1 point out of a maximum of 500. The secondary metric gives an impression of the generalizability of current text-based adventure AIs, and can be used as a tie breaker in case two or more agents perform equally well according to the first metric. In case of two agents performing the same on both metrics (which has not happened so far), we would use an additional tie breaking criterion, preferring the agent that is using the least prior domain knowledge of text-based adventure games.

\section{RESULTS\label{sec:results}}

\begin{table}
\centering
\begin{tabular}{@{}lrrrrr@{}}
\toprule
Agent & \multicolumn{2}{c}{\% completion} & \multicolumn{2}{c}{\% non-zero} \\
\cmidrule{2-3} \cmidrule{4-5}
& M & SD & M & SD \\
\midrule
\textsc{BYU\-Agent 2016} & 0.79 & - & 15 & - \\
\textsc{Golovin} & 1.45 & 0.09 & 31 & 3.94 \\
CARL (\textsc{BYU\-Agent 2017}) & 1.59 & - & 30 & - \\
NAIL & {\bf 2.56} & 0.33 & {\bf 45.5} & 2.84 \\
\midrule
\textsc{Golovin} (100 steps) & 0.99 & 0.24 & 17.5 & 3.53 \\
NAIL (100 steps) & 0.95 & 0.19 & 26 & 2.11 \\
\textsc{Golovin} (10k steps) & 1.44 & 0.10 & 32.5 & 4.25 \\
\midrule
\textit{RandomAgent} & 1.66 & 0.15 & 34 & 2.11 \\
\bottomrule
\end{tabular}
\caption{Performance on the test set of 20 games in (unless stated otherwise) 1000 time steps per game. ``\% completion'' is the average score percentage an agent achieved over all games and runs; ``\% non-zero'' is the percentage of games in which an agent achieved any score, averaged over all runs. Standard deviations (SD), whereever given, refer to 10 runs over all games. Where they are not given, only 1 run could be completed.}
\label{agentperformance}
\end{table}


The \textsc{BYU\-Agent 2016} was the strongest agent in our first competition in 2016. In 2017, CARL was the strongest agent, improving on both \textsc{BYU\-Agent 2016} and \textsc{Golovin} on the primary metric. As of 2018, NAIL is the strongest text-based adventure game playing agent. It is a clear improvement over all previously submitted agents in both evaluation metrics.

Our new testing framework also enables us to conduct more in-depth comparisons of different agents' performance over time. Three such experiments have been done so far. The additional experiments with \textsc{Golovin} and NAIL at 100 time steps per game demonstrate that NAIL does not start out stronger than \textsc{Golovin} in the first 100 time steps (at least not with respect to the primary metric), but makes more effective use of additional time when scaling up to the 1000 time steps required by the competition. Additionally, the experiment running \textsc{Golovin} for 10,000 time steps per game shows that NAIL and CARL are stronger than \textsc{Golovin} even if \textsc{Golovin} is given a ten-fold time advantage. This is particularly impressive considering that NAIL and CARL are arguably using less domain-specific knowledge than \textsc{Golovin}, e.g. \textsc{Golovin}'s ``battle mode'' \cite{golovin}. It is possible that this domain-specific knowledge overfits to the training set used by the authors, and generalizes less well to the games in our test set. However, NAIL and CARL do require much longer thinking time than \textsc{Golovin}, and we can confirm the competitors' claim \cite{golovin} that \textsc{Golovin} is an improvement over \textsc{BYU\-Agent 2016}. The general level of text adventure AIs has increased from every competition to the next so far.

The last row of Table \ref{agentperformance} shows the performance of our \textit{RandomAgent}, described in Section \ref{sec:competitionframework}. Only with NAIL in 2018 has the first agent managed to beat it. This is probably due to its set of pre-specified commands being extremely simple and generalizing very well across games - changing locations and trying to pick up all possible objects is typical player behavior in text-based adventure games. Again, overfitting to their own training sets might have been the problem for our competitors before NAIL, using too many time steps on larger vocabularies and more intricate strategies that do not work in as many different game situations. We are optimistic for future agents now that the \textit{RandomAgent} has been convincingly surpassed.

Considering the overall results, it is clear that all agents are facing major unsolved challenges. Even the strongest agent NAIL can currently only complete about 2.6\% of a given test game on average, and gets no points at all in more than half of them. Many of the points achieved are given by the respective game just for starting, for initially submitting the \texttt{`get all'} command in order to pick up any suitable objects in the first scene, or for walking into one of the compass directions. Despite some progress from 2016 to 2018, no agent is anywhere close to completing a single game. The problems of course include the very difficult scientific questions of how to extract information from natural language text, and how to map it to effective action in turn described in natural language, possibly via an intermediary model of the agent's environment. For example, the \textsc{BYU\-Agent 2016} is unable to deal with games that require giving prepositional commands \cite{byu2016} such as \texttt{`give dagger to wizard'}, or inferring the correct term for manipulable objects (such as requiring the command \texttt{`get shiny object'} after describing ``something shiny''). 

Looking at the design of all agents, they appear to share a common assumption that all actions which lead to new states are beneficial. Therefore these agents are essentially overcoming the challenge of sparse rewards from the environment by following an innate behaviour akin to work on curiosity as an intrinsic motivation \cite{pathak2017curiosity} or novelty search \cite{lehman2008exploiting}. Whilst these approaches have shown broad applicability, it is feasible to construct a pathological text-adventure game where such strong exploration would be punished, and an agent only rewarded for reaching a small subset of states. In the future hopefully more goal-oriented agents will be developed, such that good and bad state changes can be distinguished and maybe even predicted.

From analysing the gameplay logs we can also conclude that more technical challenges are common to most agents as well, namely the correct parsing of game output, and differentiating between in-game and out-of-game messages to the player. All games conform to the type of narration-action loop typical for text adventure games; but as they are originally written for human players, many games require a greater amount of flexibility than currently supported by AI agents. As an example, many games begin with an out-of-game message such as \texttt{`Would you like to resume a saved game (Y/N)?'}. Reacting to these as if they were in-game narratives often leads to complete failure at the game. Some games respond to the input \texttt{`hint'} with an in-game hint regarding the puzzle at hand, from which the agents can successfully retrieve information such as nouns and verbs to try in future actions; however, some games respond to \texttt{`hint'} by opening an out-of-game multiple-choice menu, which agents currently cannot handle. Furthermore, some agents have trouble identifying the score they have achieved, because they are expecting it in a slightly different format from what the game at hand returns. The game could for example respond to the input \texttt{`score'} with \texttt{`If you were to stop now, you would score 50 points out of a maximum of 1500'}, instead of the expected \texttt{`You have so far scored 50 out of a possible 1500, in 23 turns'}. \textsc{Golovin} for example stops playing if it cannot identify the game score for 10 successive attempts -- however, it aborts many games with valid but unexpected score formats, while playing others that actually do not keep a score but return a message in the expected format because the author did not care to remove this functionality (e.g. \texttt{`You have so far scored 0 out of a possible 0, in 23 turns'}).

The creativity of interactive fiction writers can lead to even greater challenges in parsing and/or scoring, but along with scoreless games these challenges have been removed from the test set for the foreseeable future. For example, games can have a non-numerical ranking (``captain'') instead of a score; they can use made-up words or languages (even for reporting the score); or they can only reward points after the player has finished significant parts of the game, giving no clear indication that the current score is zero before those milestones are reached.

Additionally, as the 2016 and 2017 competitions were not originally planned to use a test set of multiple games, we did not require functionality from the participants that would make (repeated) testing on such a set feasible or convenient. While \textsc{Golovin} for example recovered gracefully from failed interactions with individual games by reporting a result of zero points, the BYU agents frequently did not start at all on a given game, or even froze the OS. In combination with parsing and scoring problems, this made it necessary to manually re-start and supervise the agents on each game, which is why we can only report the result from a single run per test game in Table \ref{agentperformance}. In future competitions, we will improve our descriptions of the necessary functionality, probably setting time limits per time step as well, to allow for a more streamlined, fair, and reliable agent evaluation.

Finally, we note that the published version of \textsc{BYU\-Agent 2016} \cite{byu2016} as well as some prior work \cite{DBLP:journals/corr/NarasimhanKB15} use reinforcement learning, playing a given game many times and gradually improving performance by learning successful commands for the game at hand. So far however, our competition has not provided a learning track, and therefore required the submission of fully trained agents, or agents capable of one-shot learning within a single game-playing episode. This does not fully reflect the capabilities of some agents. Therefore, we are considering the introduction of a learning track. In addition to the improvements for future competitions described above, this track could profit from the lessons learned by the broader reinforcement learning community on the topic of evaluating and comparing agents \cite{DBLP:conf/aaai/0002IBPPM18,machado2017revisiting}. For example, (1) reporting agent's average performance at several fixed stages of training to enable comparison of both final performance and rate of learning; (2) running multiple repeats to evaluate variance in learning performance due to the known issues of robustness and reproducibility with modern reinforcement learning algorithms; and (3) requiring all entrants to open source agent submissions and fully document hyper-parameter settings. 

\section{CONCLUSION\label{sec:conclusion}}
Contemporary machine learning techniques have recently had many successes in game-playing domains such as Go that are traditionally hard for AI \cite{silver2017mastering}. While it is clear that games provide an artificially restricted domain, we claim in this article that there is a tendency for game AI domains to be chosen such that the applicable operators are known in advance. This bias effectively means that the \emph{epistemological} problems of AI, first raised by John McCarthy \cite{McCarthy:1977:EPA:1622943.1623044,McCarthy:1987:PPS:42641.42642}, are neglected. Broadly, such problems are concerned with extracting salient knowledge in non-trivial environments. This is unfortunate, since they are highly relevant for many real-world applications of natural language processing: The BYU team, for example, has recently acquired funding from Amazon as part of the ``Alexa Prize'', a challenge to create social bots that can converse coherently and engagingly with humans. They informed us ``Some of the research that went into CARL was foundational in our approach to creating [the Alexa competitor] EVE. So it might please you to know that the CIG competition is having ripples with pretty wide impact'' (Nancy Fulda, personal communication, Feb. 6, 2018).

The Text-Based Adventure AI Competition arose out of a desire to re-emphasise these neglected aspects of AI and motivate further experimentation. To date, mostly model-free approaches have been used as predicting the state transitions that might be caused by future actions proves to be very challenging. \textsc{Golovin}'s exploration command generator and NAIL's knowledge graph have made some first steps, but there is still much room for improvement, and we are far from a consensus on how to optimally tackle the related problems. We hope that the community will perceive the challenge offered by this competition as both long-standing and meaningful, and respond with new  approaches that push the envelope of existing wisdom regarding both hard problems and good solution mechanisms.


\section*{ACKNOWLEDGMENTS}
We would like to acknowledge the hard work of all competitors. Specifically, our thanks go to Nancy Fulda, Daniel Ricks, Ben Murdoch, and David Wingate from Brigham Young University, USA; Bartosz Kostka, Jaros\l{}aw Kwiecie\'n, Jakub Kowalski, and Pawel Rychlikowski from the University of Wroc\l{}aw, Poland; and Matthew Hausknecht, Ricky Loynd, Shuohang Wang, and Greg Yang from Microsoft Research. We would also like to thank Karthik Narasimhan and Tejas D. Kulkarni for their input when originally defining the format of the first competition.


\addtolength{\textheight}{-12cm}   



\bibliographystyle{IEEEtran}
\bibliography{artificial-adventurer}

\begin{thebibliography}{10}
\providecommand{\url}[1]{#1}
\csname url@samestyle\endcsname
\providecommand{\newblock}{\relax}
\providecommand{\bibinfo}[2]{#2}
\providecommand{\BIBentrySTDinterwordspacing}{\spaceskip=0pt\relax}
\providecommand{\BIBentryALTinterwordstretchfactor}{4}
\providecommand{\BIBentryALTinterwordspacing}{\spaceskip=\fontdimen2\font plus
\BIBentryALTinterwordstretchfactor\fontdimen3\font minus
  \fontdimen4\font\relax}
\providecommand{\BIBforeignlanguage}[2]{{%
\expandafter\ifx\csname l@#1\endcsname\relax
\typeout{** WARNING: IEEEtran.bst: No hyphenation pattern has been}%
\typeout{** loaded for the language `#1'. Using the pattern for}%
\typeout{** the default language instead.}%
\else
\language=\csname l@#1\endcsname
\fi
#2}}
\providecommand{\BIBdecl}{\relax}
\BIBdecl

\bibitem{Montfort:2004:TLP:940352}
N.~Montfort, \emph{{Twisty Little Passages: An Approach to Interactive
  Fiction}}.\hskip 1em plus 0.5em minus 0.4em\relax MIT Press, 2004.

\bibitem{Nelson00theinform}
D.~K. Nelson, \emph{{The Inform Designer's Manual}}.\hskip 1em plus 0.5em minus
  0.4em\relax Sanderson, 2000.

\bibitem{mccarthy1963situations}
J.~McCarthy, \emph{{Situations, Actions, and Causal Laws}}, ser. Memo (Stanford
  AI Project).\hskip 1em plus 0.5em minus 0.4em\relax Comtex Scientific, 1963.

\bibitem{McCarthy:1977:EPA:1622943.1623044}
------, ``{Epistemological Problems of Artificial Intelligence},'' in
  \emph{Proceedings of the 5th International Joint Conference on Artificial
  Intelligence - Volume 2}, 1977.

\bibitem{gibson1968senses}
J.~Gibson, \emph{The Senses Considered as Perceptual Systems}.\hskip 1em plus
  0.5em minus 0.4em\relax Allen \& Unwin, 1968.

\bibitem{schank:77a}
R.~Schank and R.~Abelson, \emph{{Scripts, Plans, Goals and Understanding: An
  Inquiry into Human Knowledge Structures}}.\hskip 1em plus 0.5em minus
  0.4em\relax Hillsdale, NJ.: Lawrence Erlbaum Associates, 1977.

\bibitem{LEHNERT1981293}
W.~G. Lehnert, ``{Plot Units and Narrative Summarization},'' \emph{Cognitive
  Science}, vol.~5, no.~4, pp. 293--331, 1981.

\bibitem{ASI:ASI1}
S.~Deerwester, S.~T. Dumais, G.~W. Furnas, T.~K. Landauer, and R.~Harshman,
  ``{Indexing by Latent Semantic Analysis},'' \emph{Journal of the American
  Society for Information Science}, vol.~41, no.~6, pp. 391--407, 1990.

\bibitem{Goyal:2010:APP:1870658.1870666}
A.~Goyal, E.~Riloff, and H.~Daum{\'e}, III, ``{Automatically Producing Plot
  Unit Representations for Narrative Text},'' in \emph{Proceedings of the 2010
  Conference on Empirical Methods in Natural Language Processing}, 2010, pp.
  77--86.

\bibitem{DBLP:conf/ijcai/BranavanSB11}
S.~R.~K. Branavan, D.~Silver, and R.~Barzilay, ``{Non-Linear Monte-Carlo Search
  in Civilization II},'' in \emph{Proceedings of the 22nd International Joint
  Conference on Artificial Intelligence (IJCAI 2011)}, 2011, pp. 2404--2410.

\bibitem{riedl2012interactive}
M.~O. Riedl and V.~Bulitko, ``Interactive narrative: An intelligent systems
  approach,'' \emph{AI Magazine}, vol.~34, no.~1, p.~67, 2012.

\bibitem{hlubockyknowledge}
B.~Hlubocky and E.~Amir, ``{Knowledge-gathering Agents in Adventure Games},''
  in \emph{AAAI-04 Workshop on Challenges in Game AI}, 2004.

\bibitem{amirlearning}
E.~Amir and A.~Chang, ``Learning partially observable deterministic action
  models,'' \emph{Journal of Articial Intelligence Research}, vol.~33, pp.
  349--402, 2008.

\bibitem{DBLP:journals/corr/NarasimhanKB15}
K.~Narasimhan, T.~D. Kulkarni, and R.~Barzilay, ``{Language Understanding for
  Text-based Games Using Deep Reinforcement Learning},'' \emph{CoRR}, vol.
  abs/1506.08941, 2015.

\bibitem{zplet}
M.~T. Russotto, ``{ZPlet: A Z-Machine for Java},'' {Available online at
  \url{https://sourceforge.net/projects/zplet}}, accessed Feb 6th Feb, 2018.

\bibitem{zmachinestandardsdoc}
D.~K. Kevin~Bracey, Jason C.~Penney, ``{The Z-Machine Standards Document,
  version 1.1},'' Available at
  \url{http://inform-fiction.org/zmachine/standards/z1point1/index.html},
  February 2014, {A}ccessed 1st December 2017.

\bibitem{Russell:2003:AIM:773294}
S.~J. Russell and P.~Norvig, \emph{{Artificial Intelligence: A Modern
  Approach}}, 2nd~ed.\hskip 1em plus 0.5em minus 0.4em\relax Pearson Education,
  2003.

\bibitem{byu2016}
N.~Fulda, D.~Ricks, B.~Murdoch, and D.~Wingate, ``{What Can You Do with a Rock?
  Affordance Extraction via Word Embeddings},'' in \emph{Proceedings of the
  Twenty-Sixth International Joint Conference on Artificial Intelligence (IJCAI
  2017)}, C.~Sierra, Ed., 2017, pp. 1039--1045.

\bibitem{DBLP:journals/corr/abs-1301-3781}
T.~Mikolov, K.~Chen, G.~Corrado, and J.~Dean, ``{Efficient Estimation of Word
  Representations in Vector Space},'' \emph{CoRR}, vol. abs/1301.3781, 2013.

\bibitem{golovin}
B.~Kostka, J.~Kwiecie\'n, J.~Kowalski, and P.~Rychlikowski, ``{Text-based
  Adventures of the Golovin AI Agent},'' in \emph{Proceedings of the 2017
  {IEEE} Conference on Computational Intelligence and Games ({CIG} 2017)},
  2017, pp. 181--188.

\bibitem{DBLP:journals/jmlr/BengioDVJ03}
Y.~Bengio, R.~Ducharme, P.~Vincent, and C.~Janvin, ``{A Neural Probabilistic
  Language Model},'' \emph{Journal of Machine Learning Research}, vol.~3, pp.
  1137--1155, 2003.

\bibitem{DBLP:journals/corr/KirosZSZTUF15}
R.~Kiros, Y.~Zhu, R.~Salakhutdinov, R.~S. Zemel, A.~Torralba, R.~Urtasun, and
  S.~Fidler, ``{Skip-Thought Vectors},'' \emph{CoRR}, vol. abs/1506.06726,
  2015.

\bibitem{JoulinGBM16}
\BIBentryALTinterwordspacing
A.~Joulin, E.~Grave, P.~Bojanowski, and T.~Mikolov, ``Bag of tricks for
  efficient text classification,'' \emph{CoRR}, vol. abs/1607.01759, 2016.
  [Online]. Available: \url{http://arxiv.org/abs/1607.01759}
\BIBentrySTDinterwordspacing

\bibitem{DBLP:conf/aaai/LiebanaSTSL16}
D.~P. Liebana, S.~Samothrakis, J.~Togelius, T.~Schaul, and S.~M. Lucas,
  ``{General Video Game {AI:} Competition, Challenges and Opportunities},'' in
  \emph{Proceedings of the Thirtieth {AAAI} Conference on Artificial
  Intelligence (AAAI 2016)}, 2016, pp. 4335--4337.

\bibitem{genesereth2005general}
M.~Genesereth, N.~Love, and B.~Pell, ``{General Game Playing: Overview of the
  AAAI Competition},'' \emph{AI magazine}, vol.~26, no.~2, p.~62, 2005.

\bibitem{bellemare2013arcade}
M.~G. Bellemare, Y.~Naddaf, J.~Veness, and M.~Bowling, ``The arcade learning
  environment: An evaluation platform for general agents.'' \emph{J. Artif.
  Intell. Res.(JAIR)}, vol.~47, pp. 253--279, 2013.

\bibitem{togelius2015ai}
J.~Togelius, ``{AI researchers, Video Games are your friends!}'' in
  \emph{International Joint Conference on Computational Intelligence}.\hskip
  1em plus 0.5em minus 0.4em\relax Springer, 2015, pp. 3--18.

\bibitem{pathak2017curiosity}
D.~Pathak, P.~Agrawal, A.~A. Efros, and T.~Darrell, ``Curiosity-driven
  exploration by self-supervised prediction,'' in \emph{International
  Conference on Machine Learning}, 2017, pp. 2778--2787.

\bibitem{lehman2008exploiting}
J.~Lehman and K.~O. Stanley, ``Exploiting open-endedness to solve problems
  through the search for novelty,'' in \emph{ALIFE}, 2008, pp. 329--336.

\bibitem{DBLP:conf/aaai/0002IBPPM18}
P.~Henderson, R.~Islam, P.~Bachman, J.~Pineau, D.~Precup, and D.~Meger, ``{Deep
  Reinforcement Learning That Matters},'' in \emph{Proceedings of the
  Thirty-Second {AAAI} Conference on Artificial Intelligence (AAAI-18)}, S.~A.
  McIlraith and K.~Q. Weinberger, Eds.\hskip 1em plus 0.5em minus 0.4em\relax
  {AAAI} Press, 2018, pp. 3207--3214.

\bibitem{machado2017revisiting}
M.~C. Machado, M.~G. Bellemare, E.~Talvitie, J.~Veness, M.~Hausknecht, and
  M.~Bowling, ``{Revisiting the Arcade Learning Environment: Evaluation
  Protocols and Open Problems for General Agents},'' \emph{arXiv preprint
  arXiv:1709.06009}, 2017.

\bibitem{silver2017mastering}
D.~Silver, J.~Schrittwieser, K.~Simonyan, I.~Antonoglou, A.~Huang, A.~Guez,
  T.~Hubert, L.~Baker, M.~Lai, A.~Bolton \emph{et~al.}, ``{Mastering the Game
  of Go without Human Knowledge},'' \emph{Nature}, vol. 550, no. 7676, pp.
  354--359, 2017.

\bibitem{McCarthy:1987:PPS:42641.42642}
J.~McCarthy and P.~J. Hayes, ``{Readings in Nonmonotonic Reasoning},'' M.~L.
  Ginsberg, Ed.\hskip 1em plus 0.5em minus 0.4em\relax Morgan Kaufmann
  Publishers Inc., 1987, ch. {Some Philosophical Problems from the Standpoint
  of Artificial Intelligence}, pp. 26--45.

\end{thebibliography}

\end{document}